\documentclass[abstract]{scrartcl}
\usepackage[utf8]{inputenc}
\usepackage[T1]{fontenc}
\usepackage[english]{babel}
\usepackage[autostyle=true]{csquotes}
\usepackage{amsmath}
\usepackage{amsfonts}
\usepackage{amssymb}
\usepackage{amsthm}
\usepackage{microtype}
\usepackage{multicol}
\usepackage{lmodern}
\usepackage{mathtools}
\usepackage[colorlinks,pdfpagelabels,pdfstartview = FitH, bookmarksopen = true,bookmarksnumbered = true, linkcolor = black, plainpages = false, hypertexnames = false, citecolor = black]{hyperref}
\usepackage{bm}
\usepackage[shortlabels]{enumitem}
\usepackage{booktabs}
\usepackage{subcaption}
\usepackage{authblk}

\usepackage[square,numbers, sort]{natbib}

\newcommand{\subheader}{\noindent\textbf}
\newcommand{\R}{\ensuremath{\mathbb{R}}}

\renewcommand{\phi}{\varphi}
\renewcommand{\theta}{\vartheta}
\newcommand{\ie}{\mbox{i.e.}}
\newcommand{\eg}{\mbox{e.g.}}

\title{The Effects of Randomness on the Stability of Node Embeddings}

\author{Tobias Schumacher}
\author{Hinrikus Wolf}
\author{Martin Ritzert}
\author{Florian Lemmerich}
\author{Jan Bachmann}
\author{Florian Frantzen}
\author{Max Klabunde}
\author{Martin Grohe}
\author{Markus Strohmaier}
\affil{RWTH Aachen University}

\begin{document}
\maketitle
\begin{abstract}
  We systematically evaluate the (in-)stability of state-of-the-art node embedding algorithms due to randomness,
  i.e., the random variation of their outcomes given identical algorithms and graphs. 
  We apply five node embeddings algorithms---HOPE, LINE, node2vec, SDNE, and GraphSAGE---to synthetic and empirical graphs and assess their stability under randomness with respect to (i) the geometry of embedding spaces as well as (ii) their performance in downstream tasks.
  We find significant instabilities in the geometry of embedding spaces independent of the centrality of a node. 
  In the evaluation of downstream tasks, we find that the accuracy of node classification seems to be unaffected by random seeding while the actual classification of nodes can vary significantly.
  This suggests that instability effects need to be taken into account when working with node embeddings.
  Our work is relevant for researchers and engineers interested in the effectiveness, reliability, and reproducibility of node embedding approaches.
  
\end{abstract}

\begin{figure}
\centering
	\includegraphics[width=.9\columnwidth]{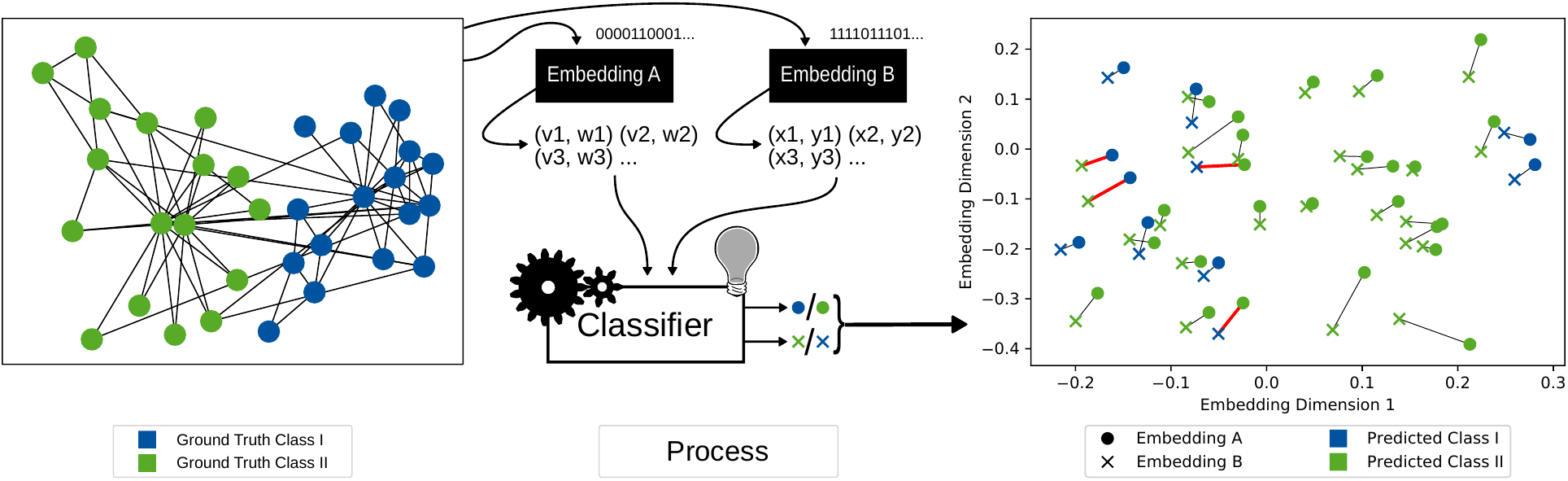}
	\caption{\emph{Illustration}.
	Stability of node embeddings with respect to their (i) geometry and (ii) predictive (classification) power.
    Left: Zachary's Karate Club Graph \cite{zachary1977information}
Middle: Comparison approach.
    Right: Geometric and predictive stability, illustrated with a decision tree classifier. Lines are used to visually connect identical nodes between embeddings A and B.
} \label{fig:motivation}
\end{figure}

\section{Introduction}\label{sec:introduction}

Many state-of-the-art node embedding algorithms make explicit use of randomness in parameter initialization, edge sampling, or through stochastic optimization.
Thus, the application of the same algorithm with the same parameters on the exact same graph data can lead to different embeddings as illustrated in Figure \ref{fig:motivation}.

\subheader{Research objective. }
We investigate the effects of randomness on the stability of node embeddings.
Towards this end, we assess (i) \emph{geometric stability} of node embeddings, \ie{}, changes in the embedding spaces  and (ii) \emph{downstream stability}, \ie{}, the stability of node classification outcomes.
Understanding the effects of randomness on the stability is a prerequisite for more reliable and reproducible deployments of node embedding techniques in the future.

\subheader{Approach.}
We conduct experiments on synthetic and real-world graph data. 
For five state-of-the-art algorithms, we compute multiple node embeddings with the same parameters on the same data but with different random seeds.
Specifically, we apply HOPE \cite{ou2016hope}, LINE \citep{tang2015line}, node2vec \cite{grover2016node2vec}, SDNE \citep{wang2016sdne}, and GraphSAGE~\cite{hamilton2017graphsage}.
First, we compare the outcomes of the individual runs with respect to measures capturing the geometry in the embeddings space (e.g., nearest neighbors of a node) and analyze the influence of graph properties and node centrality on stability.
Second, we quantify stability with respect to downstream tasks, in particular node classification.

\subheader{Results and implications.}
We find substantial geometric instabilities for most node embedding algorithms. 
An exception to this is HOPE, which produces nearly constant embeddings.
We show that downstream node classification accuracy seems robust.
Nevertheless, the actual predicted classes of individual nodes often differ between classifiers trained on different embeddings. 
Together with possible extensions on the effects of structural noise or modifications in the input graphs, this work paves the way towards a more fundamental understanding of the general reliability of node embeddings.

\section{Stability Measures}\label{sec:stabilityMeasures}
We define stability of an embedding algorithm informally as the difference between two embeddings due to randomness in the embedding algorithm.
This section provides some basic notations and discusses measures for quantifying the stability of embedding algorithms. 
We categorize these measures into two groups:
measures based on the geometric stability in the embedding space, and measures based on outcomes of downstream tasks.
The latter are tasks that utilize node embeddings to accomplish a specific machine learning task.
As representative downstream tasks, we have chosen typical node classification problems.

\subsection{Basic Definitions} 
  A graph $G=(V,E)$ is given by the set of vertices $V = \{v_1,\dots, v_N\}$ and edges $E$.
We define a node embedding as a function \mbox{$\varphi \colon V \rightarrow \R^d$}
  which maps nodes to embedding vectors of length $d\in N$.
  In our experiments, we produce multiple embeddings for each graph and thus use upper indices $\varphi^{(l)}(v_i) = z_i^{(l)} , \varphi^{(m)}(v_i) = z_i^{(m)}$ to indicate different embeddings, both on the functions as well as the embedding vectors.
  We also denote a node embedding $\varphi^{(l)}(V)$ by its embedding matrix $Z^{(l)} = \big[z^{(l)}_1, \dots, z^{(l)}_N\big]^T\in\R^{N\times d}$
  in which the $i$-th row corresponds to the embedding $z_i^{(l)}$ of node $v_i$ under $\varphi^{(l)}$.

\subsection{Geometric Stability}

    First, we discuss geometric stability measures which are inspired by related literature on word embeddings~\cite{hellrich2016bad, antoniak2018stability, hamilton2016diachronic, hamilton2016cultural}.
Finding a measure is not trivial since the outcome of embedding algorithm runs might result, \eg{}, in rotations of one another.
In this case the respective embedding matrices differ substantially from each other but the embeddings are equivalent in essence and effect.
As a consequence, simple and established similarity measures such as the Frobenius distance between two embedding matrices are not suitable.
    Further, $L_p$-norm based measures such as Euclidean distance are known to suffer from the curse of dimensionality, resulting in large distances between almost every pair of vectors.
    For that reason, we decided on three angle-based measures.
    All these measures utilize the cosine similarity $\operatorname{cos-sim}(v,w):=\langle v,w\rangle/\|v\|\|w\|$, which gives the cosine of the angle between two vectors $v,w\in\mathbb{R}^n$.
Every measure computes a score for a single node in two embeddings. 
    In order to obtain a score for an embedding space to compare different algorithms, we average over all pairs of embeddings and all nodes.

\subheader{Aligned Cosine Similarity\cite{hamilton2016diachronic}.}
    This measure computes the node-wise cosine similarity between two embeddings after aligning the axes of the corresponding embedding spaces.
    To obtain the optimal alignment, we normalize all embedding vectors and solve the Procrustes problem:
    Given two embedding matrices $Z^{(l)}, Z^{(m)}$, determine the transformation matrix $Q^{(l,m)}\in\R^{d \times d}$ by solving the minimization problem
\[
         Q^{(l,m)} \coloneqq \underset{Q^TQ=I}
         {\operatorname{argmin}}
         \left\|Z^{(l)} Q -  Z^{(m)}\right\|_F .
    \]
    Using $Q^{(l,m)}$, we define the \emph{aligned cosine similarity} between the embeddings $z_i^{(l)}, z_i^{(m)}$ of a node $v_i$ by the cosine similarity between two embeddings of node $v_i$ after aligning the embedding spaces:
    \begin{equation*}\label{eq:acos}
      s^\text{acs}\left(z_i^{(l)}, z_i^{(m)}\right) \coloneqq
      \operatorname{cos-sim}\left( z_i^{(l)} Q^{(l,m)},\, z_i^{(m)}\right).
    \end{equation*}We also experimented with centering the embedding spaces before applying the rotation. 
    This led to decreased similarity values for LINE and node2vec which might be due to the algorithm's use of the origin as an anchor during computation.
    For the other algorithms, centering did not affect the outcome significantly.

\subheader{$\bm{k}$-NN Jaccard Similarity.}
This measure is based on a local neighborhood perspective.
    In both embedding spaces, we compute for a node $u$ the $k$ nearest neighbors with respect to cosine similarity.
    We then calculate the Jaccard similarity of the two nearest-neighbor sets of \(u\).
    Formally, let $\mathcal{N}_k^{(l)}(v_i)$ denote the set of the $k$ nearest neighbors $\{ v_{n_1},\dots, v_{n_k} \}$ between the corresponding embedding vectors $z^{(l)}_i$ and $z^{(l)}_{n_j}$ with respect to aligned cosine similarity.
    The \emph{$k$-NN Jaccard similarity} between two embeddings $z_i^{(l)}$ and $z_i^{(m)}$ of a node $v_i$ is then given by 
    \begin{equation*}\label{eq:knn}
        s_k^\text{NN}\left(z_i^{(l)}, z_i^{(m)}\right) \coloneqq \tfrac{|\mathcal{N}_k^{(l)}(v_i)\,\cap\, \mathcal{N}_k^{(m)}(v_i)|}{|\mathcal{N}_k^{(l)}(v_i)\,\cup\, \mathcal{N}_k^{(m)}(v_i)|}.
    \end{equation*}

\subheader{Second-Order Cosine Similarity \cite{hamilton2016cultural}.}
    This measure also evaluates the local neighborhood structure, but in a more fine-grained way.
    Formally, we consider the ordered set $\{u_1,\dots, u_K\}\coloneqq \mathcal{N}_k^{(l)}(v_i)\cup \mathcal{N}_k^{(m)}(v_i)$ of the $k$ nearest neighbors of $v_i$ in each of the two embeddings.
    On both embeddings $\varphi^{(l)}(V)$, $\varphi^{(m)}(V)$ and for each $u_j$, we compute the cosine similarity of its embedding with the respective embedding of $v_i$.
    Next, for all $v_i$ we store the resulting values in vectors $s^{(l)}(v_i),s^{(m)}(v_i)$, where 
    \[
        s^{(l)}_j\left(v_i\right)  \coloneqq \operatorname{cos-sim}\left(z_i^{(l)}, \varphi^{(l)}(u_j)\right) \quad\forall u_j\in \mathcal{N}_k^{(l)}(v_i)\cup \mathcal{N}_k^{(m)}(v_i).
    \]
    Then, the \emph{second-order cosine similarity} between the embeddings $z_i^{(l)}, z_i^{(m)}$ of $v_i$ with respect to neighborhood size $k$ is defined as
    \begin{equation*}\label{eq:2ndcos}
    s^\text{cos}_k\left(z_i^{(l)}, z_i^{(m)}\right) \coloneqq  \operatorname{cos-sim}\left(s^{(l)}_j(v_i),\,s^{(m)}_j(v_i)\right).
    \end{equation*}
    Intuitively, this measure quantifies the similarity of the distances of a node to its nearest neighbors between two embeddings.

\subsection{Downstream Stability} \label{sec:downstream_measures}
Instability of node embeddings potentially induces instability in the results of classifiers based on these embeddings.
We refer to this as \emph{downstream (in)stability}.
To quantify the downstream stability, we train (i) multiple classifiers on the same embedding and (ii) multiple classifiers on multiple embeddings of the same graph produced by the same embedding algorithm.
Differences in the classifications in~(i) indicate the stability of the classification algorithm itself due to random elements in the classification algorithm, independent of the embedding.
Comparing outcomes of classifiers trained with different embeddings (ii) provides indication for the combined stability of the embedding algorithm and the classifier.
Thus, the difference between outcomes of (i) and (ii) corresponds to the influence of the instability of the embeddings on the stability of the classification. 
To measure differences in the outcome of classifiers, we compute general performance scores (such as micro-F1 of the classification on a holdout set) as well as the \emph{stable core}, \ie{}, the ratio of nodes that are assigned to the same class by multiple classifier runs.

\section{Experimental Framework}\label{sec:experimentalFramework}

Our experiments quantify the stability of five state-of-the-art node embedding algorithms.
We start with a short description of the algorithms and datasets and then describe the experiments.

\subsection{Node Embedding Algorithms}\label{subsec:algorithms}
We consider five state-of-the-art node embedding algorithms as representatives of the spectrum of currently existing approaches.
We provide the key ideas of the algorithms and explicitly point out the use of randomness.
    
    \subheader{HOPE.} 
        The spectral node embedding algorithm HOPE \cite{ou2016hope} obtains its embeddings from factorizing the Katz similarity matrix.
        For this decomposition, a generalized SVD approximation is utilized.
The random initialization in the SVD approximation is the only source of randomness in this algorithm.
        Essentially, this makes HOPE an almost deterministic embedding algorithm.

    \subheader{LINE.}
        The embedding method LINE \citep{tang2015line} concatenates two embeddings which preserve first and second-order proximities respectively.
The second-order proximity is optimized via skip-gram utilizing context vectors, negative sampling and weight-based edge sampling.
In total, the random initialization and optimization of embedding vectors, both sampling approaches, and a densification of the input graph in a preprocessing step yield four major sources of randomness in this algorithm.

    \subheader{node2vec.} 
        The random walk based approach node2vec \cite{grover2016node2vec} is an embedding algorithm on nodes inspired by word2vec \cite{mikolov2013distributed} and DeepWalk \cite{perozzi2014deepwalk}.
        Node2vec applies random walks over the input graph to determine which nodes appear in the same context. 
        These contexts are then fed into the skip-gram model with negative sampling.
        Next to the random walks and negative sampling, the initialization and optimization of the embedding vectors are sources of randomness.

    \subheader{SDNE.}
    	Structural deep network embedding (SDNE) jointly optimizes a local and global similarity measure on weighted graphs by utilizing a semi-supervised neural autoencoder \citep{wang2016sdne}.
    	The model is optimized using SGD, and thus, the initialization and optimization of the weights introduced by the layers of the autoencoder account for numerous random processes.

    \subheader{GraphSAGE.} 
        The embedding algorithm GraphSAGE \cite{hamilton2017graphsage} applies a GNN to compute its embeddings.
        First, the GNN's aggregation functions for the neighborhood are trained.
        Then, the embeddings are computed by applying the GNN to the initial embeddings (i.e. the labels) of the nodes instead of optimizing the embeddings directly.
GraphSAGE samples a fixed-size neighborhood for each node over which the aggregation is performed. 
        This sampling accounts for the main randomness in the algorithm.

\subsection{Graph Datasets}\label{sec:data}
 
  We investigate the stability of node embeddings on both empirical and synthetic graphs. 
  While empirical graphs provide labels which we can use to analyze downstream stability, synthetic graphs allow us to measure the effect of varying graph size and density on the geometric stability of the embeddings.

\subheader{Synthetic graphs. }
  We utilize two graph models, namely Barabasi-Albert \cite{barabasi1999emergence} and Watts-Strogatz \cite{watts1998collective} graphs.
For each model we generate two sets of graphs, in which we either fixed the graph's size and varied its density, or vice versa.
  The first set consists of nine graphs with $8000$ nodes each and densities $D\in \{0.00025, 0.0005, 0.001,$ $0.002, 0.005,$ $0.01, 0.02, 0.05, 0.1 \}$.
  The second set consists of seven graphs with a varying size of $N\in\{2^k\cdot 10^3 , k\in\{0,1,\dots, 6\} \}$ nodes and a fixed density $D=0.01$, which resembles the sparsity that is commonly found in empirical graphs. For Watts-Strogatz graphs, we use a rewiring probability of $0.1$.
  
  \subheader{Empirical graphs. }
  We consider a broad spectrum of commonly-used graph types, including biological graphs, social networks and citation networks.
  Statistics for each graph can be found in Table \ref{tab:graphstats}.
\begin{table}[b]
  	\caption{\textit{Statistics of empirical graph datasets. } We show number of nodes (|V|) and edges (|E|), density, and number of node labels. MC indicates multi class, ML multi label problems.}
    \centering
    \begin{tabular}{lcccc}
      \toprule
      Data Set &  $|V|$ & $|E|$ & Density & \# Labels \\  \midrule
      BlogCatalog & 10,312 & 333,983 & 0.00628 & 39 (ML) \\
      Cora & 23,166 & 91,500 & 0.00034 & 10 (MC) \\
      Facebook & 7,057 & 89,455 & 0.00359 & - \\
      Protein & 3,890 & 76,584 & 0.01012 & 50 (ML) \\
      Wikipedia & 4,777 & 184,812 & 0.01620 & 40 (ML) \\\bottomrule
    \end{tabular}
    \label{tab:graphstats}
  \end{table}
\begin{itemize} [leftmargin=*]
  	\item \textbf{BlogCatalog}: This graph models the relationships among the users of the BlogCatalog website.
      Each user is represented by a node and two nodes are connected if the respective users are friends.
      Each user additionally has one or more labels which correspond to the news category their blog belongs to.
  	\item \textbf{Cora} \citep{subelj2013cora}:
      In the well-known Cora citation network
each scientific paper is represented by a node, and a directed edge indicates that the outgoing node cites the target node. 
      Each paper is associated with a category that refers to its research topic.
  	\item \textbf{Facebook} \citep{rozemberczki2018gemsec}:
      The Facebook government dataset models the social network structure of verified government sites on Facebook.
      Thus, each site is represented by a node and nodes are connected by an edge if both sites like each other.
  	\item \textbf{Protein} \citep{stark2006biogrid}:
      This biological network models protein interactions in human beings.
      Each node represents a protein and two nodes are connected if the corresponding proteins interact with each other.
      Additionally, each node is associated with one or more labels that represent biological states.
  	\item \textbf{Wikipedia} \citep{mahoney2011text}:
      This network represents the co-occurrence of words within a dump of Wikipedia articles.
      Each word corresponds to a node, and weighted edges represent the number of times two words occur in the same context.
      Additionally, each node has one or more labels that encode its part of speech. 
  \end{itemize}
We used the Cora dataset from the KONECT graph repository \cite{Kunegis2013konect} and BlogCatalog from the ASU Social computing repository \cite{zafarani_asu}.
  The other empirical datasets were taken from the SNAP graph repository \cite{leskovec2014snap}.

\subsection{Experimental Setup}\label{subsec:setup}
  For every algorithm, we compute $30$ embeddings of each dataset.
  We use the embedding dimension $d=128$ for all algorithms.
  To analyze the stability of the embedding algorithms, we apply the presented geometric and downstream measures.
  For every algorithm from Section~\ref{subsec:algorithms} we use the reference implementation except for HOPE, for which no reference implementation was published.
  Thus we resorted to the HOPE implementation from the GEM library \cite{Goyal2018GEM}.
  We run the algorithms with default parameters from the given implementations whenever possible.
  We adapted SDNE to use only a single intermediate layer and for larger graphs increased the weight on the reconstruction error and the regularization term, as otherwise SDNE maps all nodes onto the same vector.
  The code for our experiments is published on GitHub.\footnote{All code available on \url{https://github.com/SGDE2020/embedding_stability}}

\subheader{Geometric Stability. }
  For the described geometric measures we compute the node-wise similarity values of the embeddings.
  We use $k=20$ for the neighborhood size of the $k$-NN Jaccard similarity and the second-order cosine similarity.
  This value provides an adequate trade-off between the noise from considering too wide or too narrow neighborhoods, as tested in a preliminary experiment.
Additionally, we examine the correlation of node embedding instability with centrality of the respective node in the graph as measured by degree, betweenness centrality, coreness, or Page\-Rank.
  In the results described here, we focus on PageRank centrality as other centrality measures lead to similar outcomes.
Furthermore, we analyze the influence of the distance between distinct nodes on the stability of the angles between their embeddings. 
  Finally, we utilize the synthetic graphs to evaluate the effect of varying graph size and density on the stability of their embeddings.

\subheader{Downstream Stability.}
  To analyze the influence of the stability in node embeddings on downstream tasks, we perform node classification with either multi-class or multi-label targets using AdaBoost, decision trees, random forests, and feedforward neural networks.
  
  On every embedding of the labeled datasets we performed node classification tasks (10-fold CV with 10 repetitions) with each classification algorithm.
  In node classification we predict either the class of a node, e.g., top-level research category in Cora, or a set of labels of a node, e.g., the news categories in BlogCatalog.
To evaluate classification performance, we report the micro-F1 score, noting that macro-F1 yields very similar results without adding additional insights. Regarding stability of individual predictions, we determine the stable core (cf. Section \ref{sec:downstream_measures}) of the individual predictions, and compare the instability induced from different instances of the same classifier to the instability induced from different embeddings.
  For the predictions used to compute the stable core, we used $75\%$ of the nodes of each graph for training and $25\%$ for evaluation.

For all classifiers we used the standard methods with default parameters from scikit-learn (AdaBoost, decision tree, random forest) and TensorFlow (neural networks).
  In the case of neural networks, we use a network with a single hidden layer of width $100$ with ReLu activation and an output layer with softmax or sigmoid activation depending on the classification type.
  Deeper and wider networks did not improve performance which is why we worked with this very simple architecture.

\section{Results}\label{sec:results}

In this section we present our experimental results for geometric as well as downstream stability. 
To display the distribution of a large number of values, which naturally occur when analyzing node-wise similarities or 10-fold classifications over sets of 30 embeddings, we use \emph{letter-value plots} \cite{letter-value-plot}. 
This variant of boxplots uses multiple boxes to give a more detailed illustration of the distribution of the underlying data points.

\subsection{Geometric Stability} \label{subsec:geometricStabilityResults}

\begin{figure*}
	\begin{subfigure}[b]{0.495\textwidth}
		\centering
		\includegraphics[width=\textwidth]{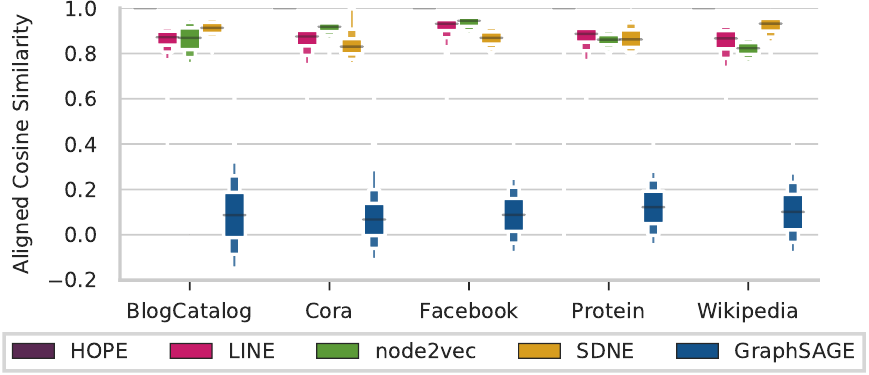}
		\caption{Variability of aligned cosine similarity $\bm{s^\text{acs}}$.}\label{fig:proc}
	\end{subfigure}
	\hfill
	\begin{subfigure}[b]{0.495\textwidth}
		\centering
		\includegraphics[width=\textwidth]{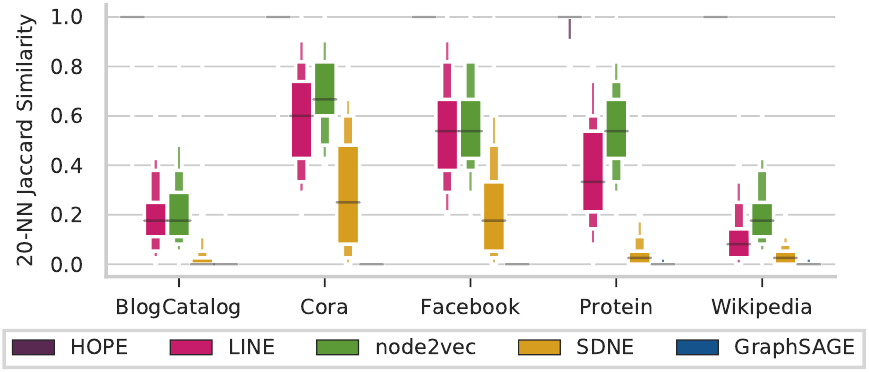}
		\caption{Variability of $\bm{k}$-NN Jaccard similarity $\bm{s_k^\text{NN}}$.}\label{fig:knn}
	\end{subfigure}
\caption{\emph{Geometric stability. } Each letter-value plot shows the node-wise similarity values resulting from 30 runs per algorithm and graph.
In (a) we use cosine similarity, in (b) we use the 20-NN Jaccard similarity. In both (a) and (b), HOPE achieves am similarity of 1.0 with hardly any variance,
whereas for GraphSAGE similarities are are around 0.1 in (a) and 0 with hardly any variance in (b).  In between, LINE, node2vec and SDNE appear moderately stable with respect to aligned cosine similarity. 
In contrast, we see in (b) that high Jaccard similarities over 0.7 rarely occur.}\label{fig:geometricMeasures}
\end{figure*}

  We start our analysis of the geometric stability by computing node-wise stability measures averaged over all pairs of embeddings.
  Figure \ref{fig:geometricMeasures} shows the distributions of (a) aligned cosine similarity and (b) $k$-NN Jaccard similarity, both over the nodes of each graph.

For the aligned cosine similarity, we observe that GraphSAGE achieves similarities that are generally only slightly above zero and sometimes even negative. 
  Negative values corresponds to angle differences of more than 90 degrees between two embeddings of the same node.
  Such angles are expected between randomly generated vectors in high dimensional spaces but not between different embedding vectors of single nodes.
HOPE yields near-constant embeddings (not shown) and shows hardly any instability.
  The algorithms SDNE, node2vec and LINE achieve aligned cosine similarities in the interval $(0.8, 0.9)$ with low variances. 
  These values correspond to angles of more than 20 degrees such that corresponding embedding vectors roughly point in the same direction after aligning the embedding spaces.
Thus, the latter algorithms exhibit a moderate, but significant degree of instability in their embeddings.

Results for the $k$-NN Jaccard similarity, as shown in Figure \ref{fig:knn}, generally confirm these findings.
  For HOPE we observe perfectly matching neighborhoods, while for GraphSAGE the neighborhoods are completely disjoint.
  This matches our observations for aligned cosine similarity.
For the other three algorithms, the resulting similarities seem to be highly dependent on the dataset with quite large variances.
Generally, node2vec appears most stable among these algorithms, though only by a slight margin over LINE.
  SDNE is the least stable embedding algorithm with respect to $k$-NN Jaccard similarity among those three algorithms, while being as stable as the other two with respect to aligned cosine similarity.
  
    \begin{figure*}
\includegraphics[width=\textwidth]{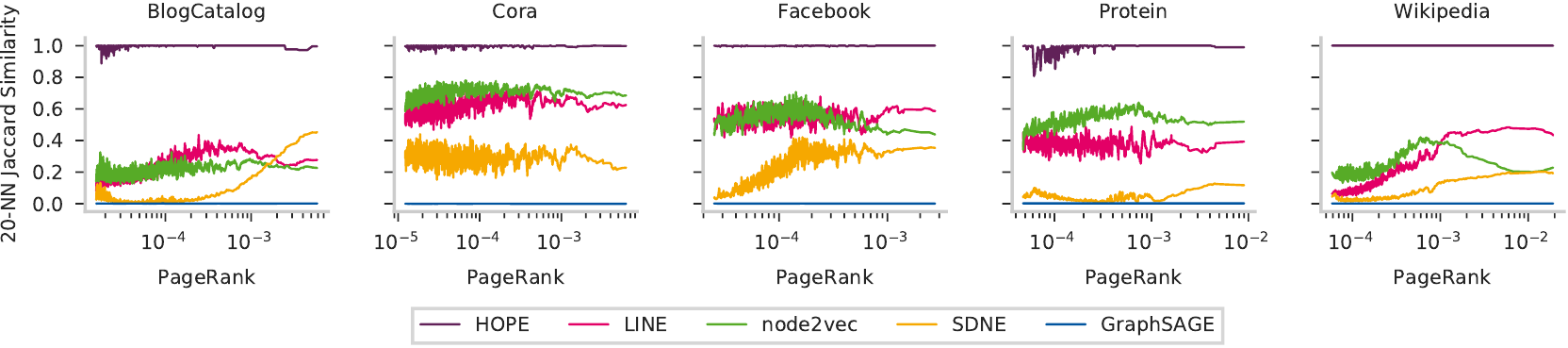}
    \caption{\emph{Influence of node centrality.} The moving average of the node-wise 20-NN Jaccard similarities resulting from 30 embeddings per graph are plotted against each node's PageRank.
    The (in-)stability of HOPE and GraphSAGE appears to be invariant to node centrality.
    For the other algorithms there is no clear correlation between the centrality of a node and the stability of its embeddings. }\label{fig:node_properties}
\end{figure*}

\begin{table}[b]
  	\caption{Average second-order cosine similarities computed over all nodes per graph and all 30 embeddings per graph and algorithm. Except for GraphSAGE, we observe outcomes close to the maximum for all algorithms on every dataset.}
    \centering
    \begin{tabular}{lccccc}
      \toprule
		Algorithm & \!\!\!\!BlogCatalog\!\!\!\! & Cora & \!Facebook\! & \!Protein\! & \!\!Wikipedia\!\! \\  \midrule
		HOPE & 1.0 & 1.0 & 1.0 & 0.9871 & 1.0 \\ 
		LINE & 0.9918 & 0.9985 & 0.9984 & 0.9909 & 0.9828 \\ 
		node2vec & 0.9974 & 0.9996 & 0.9993 & 0.9949 & 0.9895 \\ 
		SDNE & 0.9987 & 0.9981 & 0.9987 & 0.9981 & 0.9974 \\
		GraphSAGE\!\!\! & 0.1085 & 0.0414 & 0.0058 & 0.0063 & 0.0054 \\  \bottomrule

    \end{tabular}
    \label{tab:2odr-cosine}
  \end{table}
The results on second-order cosine similarity differ from the previous findings.
  In Table \ref{tab:2odr-cosine} we show the similarities per graph and embedding algorithm averaged over all nodes and embeddings. Except for GraphSAGE, which remains consistent in its low geometric stability, all embedding algorithms obtain close to the maximum node-wise second-order cosine similarities on all datasets.
  This does not contradict the observed instability of LINE, node2vec and SDNE with respect to $k$-NN Jaccard similarity. 
  Instead, these high similarity values indicate that over two embeddings, the angles of each individual node to its nearest neighbors only change marginally. 
Together with the high instabilities in the $k$-NN Jaccard similarity, this indicates that LINE, node2vec, and SDNE map similar nodes close together in the embedding space.
This might be caused by clusters where small changes in the angles between embedding vectors are able to completely change the set of nearest neighbors. A further indicator for this hypothesis are the average angles of approximately 20 degrees between embeddings of each node (cf. Figure~\ref{fig:proc}) together with the values of the second-order cosine similarity (cf. Table~\ref{tab:2odr-cosine}).

\begin{figure*}
	\begin{subfigure}[b]{0.49\textwidth}
		\centering
\includegraphics[width=\textwidth]{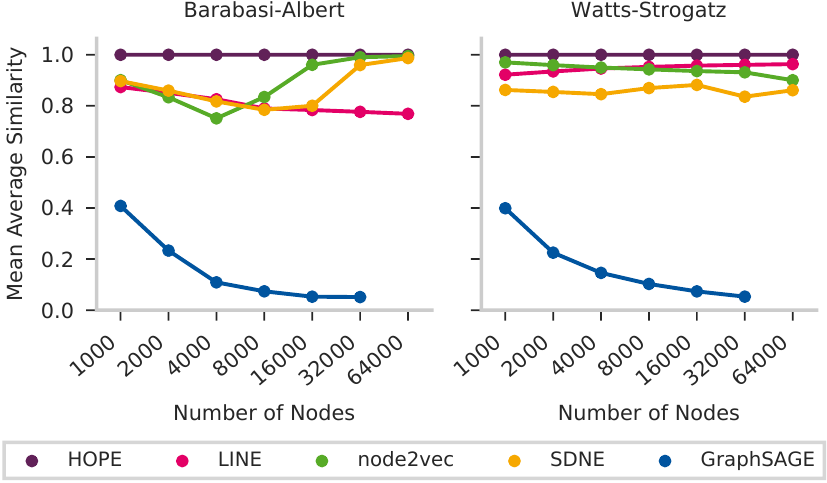}
		\caption{Mean average aligned cosine similarity over varying sizes.}\label{fig:sizes}
	\end{subfigure}
	\hfill
	\begin{subfigure}[b]{0.49\textwidth}
		\centering
		\includegraphics[width=\textwidth]{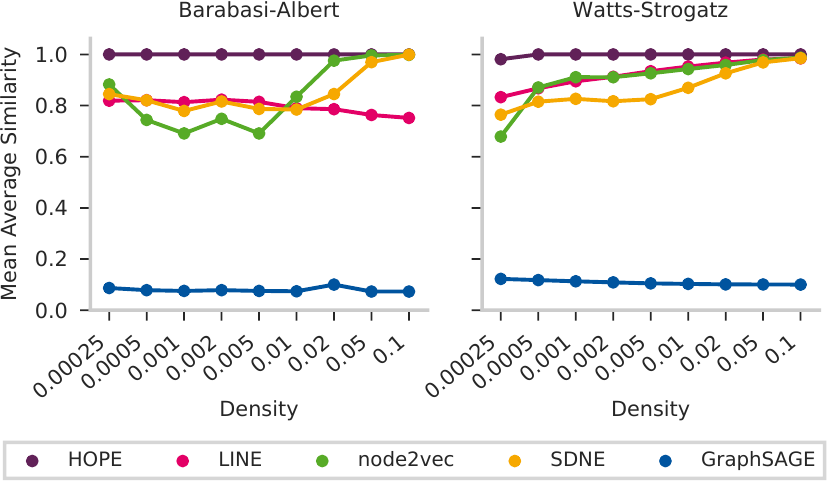}
		\caption{Mean average aligned cosine similarity over varying densities.}\label{fig:densities}
	\end{subfigure}
	\caption{\emph{Influence of graph properties.} 
In (a) synthetic graphs with varying size at fixed density 0.01 and in (b) synthetic graphs with varying density and 8000 nodes are used to measure the influence of those graph properties on stability.
Each data point represents the average node-wise similarity over all nodes per graph and all 435 embedding pairs resulting from 30 runs of the corresponding algorithm.
Except for a negative trend for GraphSAGE, the graph size does not seem to influence stability.
	Increasing density mostly leads to more stable embeddings for SDNE and node2vec, and has little or no effect on other embedding algorithms.
    }\label{fig:graph_properties}
\end{figure*}

\subheader{Influence of Node Centrality. }
  Now, we analyze whether nodes that are central in their graph have more stable embeddings.
  We expect, among others, node2vec to have more stable central nodes since these nodes occur more often in random walks.
  Thus, we plot each node's PageRank score against their average aligned cosine similarity and $k$-NN Jaccard similarity over all 30 embeddings per graph and algorithm (cf. Figure \ref{fig:node_properties}).
  We omit results with respect to aligned cosine similarity since they are very similar and do not yield any additional insights. First of all, the (in)stability of the extreme cases HOPE and GraphSAGE appears invariant of the centrality of the node.
  For LINE and node2vec there is no simple trend visible, their similarity scores look rather arbitrary.
  In particular, node2vec embeddings show no sign of increasing stability with higher centrality which does not match our expectations.
  Only for SDNE we observe a slight positive trend: on BlogCatalog, Facebook and Wikipedia stability increases with growing PageRank.

\subheader{Influence of Node Distance}
\begin{figure}
    \centering
\includegraphics{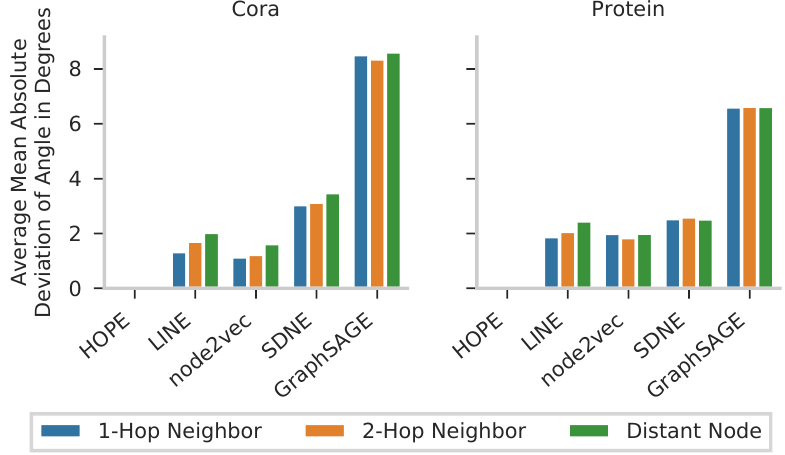}
 	\caption{\emph{Influence of node distance.} Mean absolute deviation of the angles between embedding vectors of distinct nodes over 30 embeddings. Node pairs are categorized into (i) neighboring nodes, (ii) 2-hop neighbors, (iii) more distant nodes. 1000 node pairs were sampled for each category. Except for LINE's embeddings, the variability in angles appears invariant to the distance between the corresponding nodes.} \label{fig:neighbor_meanvar}
\end{figure}
  After analyzing the stability of the embeddings of \emph{individual} nodes in previous experiments, we next study the stability of angles between the embeddings of \emph{pairs} of nodes.
  For this purpose, we measure the angles between two distinct nodes in a single embedding and compare these values across different embeddings.
  For each empirical graph, we sample 1000 pairs of 1-hop neighbors, 2-hop neighbors, and more distant nodes.
  Figure \ref{fig:neighbor_meanvar} shows the average of the distances between angles over all pairs of the 30 embeddings for each algorithm and Cora and Protein as exemplary datasets.

In line with previous results, we observe that angles are almost constant for HOPE, whereas for GraphSAGE the variances are consistently the highest by a large margin.
  In between, LINE, node2vec and SDNE display variations of mostly around two degrees or lower.
Further, the observed deviations appear mostly invariant to node distance.
  LINE behaves differently as it displays a notable trend of slightly increasing deviations with growing node distances.

\subheader{Influence of Graph Properties.}
To evaluate the impact of graph properties on the stability of the embeddings, we generated synthetic graphs with varying sizes and densities 
see Section~\ref{subsec:setup}).
In Figure~\ref{fig:graph_properties} we plot the average aligned cosine similarities over all nodes and embeddings per graph and algorithm against (a) graph size and (b) graph density. 
Figure~\ref{fig:sizes} contains missing data points that result from terminating the embedding computation after a maximum of 72 hours per embedding.

When considering the impact of the graph \emph{size}, we observe a few trends.
First of all, the stability of GraphSAGE seems to be sensitive towards graph size on both synthetic models.
For the smallest graphs, the cosine similarity has an average of roughly $0.4$, which is significantly higher than what we observed on all (larger) empirical datasets. 
With increasing graph size, this stability quickly deteriorates. 
For the other algorithms, the embeddings on Watts-Strogatz graphs show slightly better stability as in our experiments on empirical graphs.
Node2vec and LINE consistently score average aligned cosine similarities over $0.9$, and between $0.8$ and $0.9$ for SDNE.
On the Barabasi-Albert graphs we mostly observe aligned cosine similarities below \(0.9\), and diverging stability with increasing graph size.
While node2vec and SDNE perform more stable on larger graphs, LINE displays a slight downward trend.

For the dependence on graph \emph{density} plotted in Figure \ref{fig:densities}, we see that the embedding stability of SDNE and node2vec depends on the density and is maximal for the highest density graphs.
On Watts-Strogatz graphs this upward trend is continuous and also LINE follows this trend while on Barabasi-Albert graphs LINE embeddings continuously get less stable.
On Barabasi-Albert graphs of density up to $0.01$, the stability of SDNE and node2vec does not follow a clear trend.
For higher densities, i.e., graphs that are more dense than the empirical graphs, the stability goes up.
The stability of GraphSAGE is once again consistently the lowest, and appears unaffected from graph density.

In total, we see that node and graph properties as well as the distance between nodes have a rather small influence on the stability of the embeddings. 
Stability is dominated by the choice of embedding algorithms and overshadows the aforementioned effects.

\subsection{Downstream Stability}\label{subsec:downstreamResults}

To evaluate the impact of instabilities in the embeddings on downstream task, we utilize all labeled datasets. We first present results for the stability in the classification performance, before continuing with the stability in the individual predictions.

\subheader{Stability of Classification Performance.}
\begin{figure*}
\includegraphics[width=\textwidth]{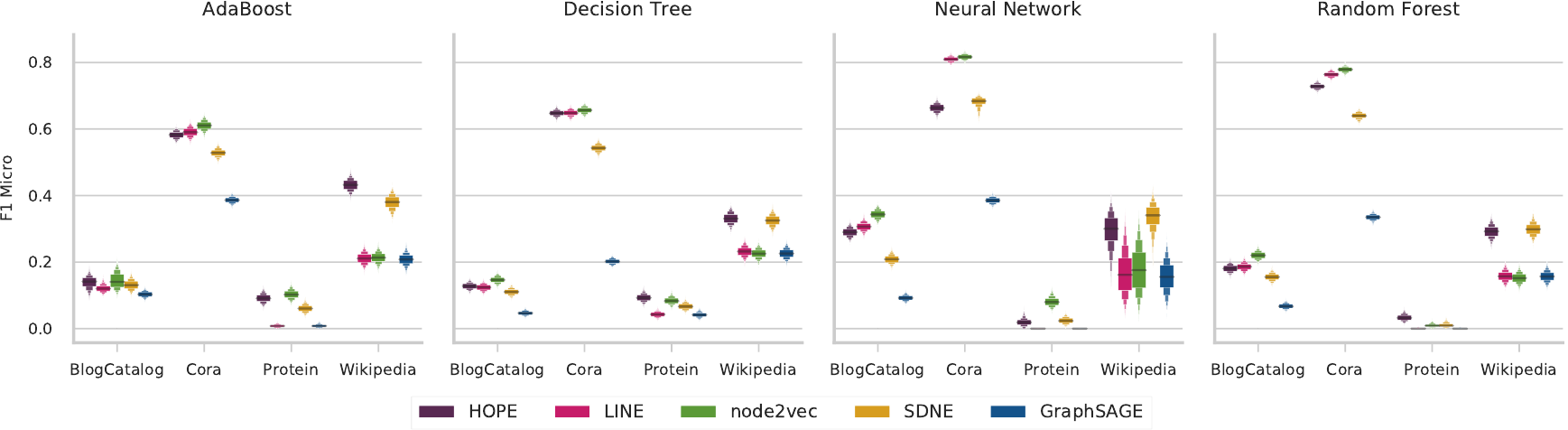}
	\caption{\emph{Stability of classification performance.} Stability of the prediction accuracies of the used classification methods plotted against the used embedding algorithms. 
		Each box corresponds to the prediction of 30 embeddings with 10-fold cross validation and 10 repetitions. In general, the classification accuracies are not influenced by the instabilities of the embeddings, since the variances in their prediction accuracies are relatively small. Higher variances appear to depend on datasets rather than embedding techniques.}
	\label{fig:acc_variability}
\end{figure*}
Figure \ref{fig:acc_variability} depicts the micro-F1 scores of the predictions on all embeddings of every dataset.
Each box in the figure aggregates the different micro-F1 scores of the repeated predictions on the 30 embedding per algorithm and dataset.
Except for neural networks on Wikipedia, we observe that the resulting F1 scores of all classification tasks vary only marginally. 
This is consistent over all embeddings methods and classifiers, which essentially appear equally stable with respect to the classification performance.
Aside from stability, we generally observe the lowest F1 scores on GraphSAGE. 
There is no clear trend regarding which embedding algorithm or classifier performs best in general.

\subheader{Stability in Node-Wise Predictions.}
\begin{figure*}
	\includegraphics[width=\textwidth]{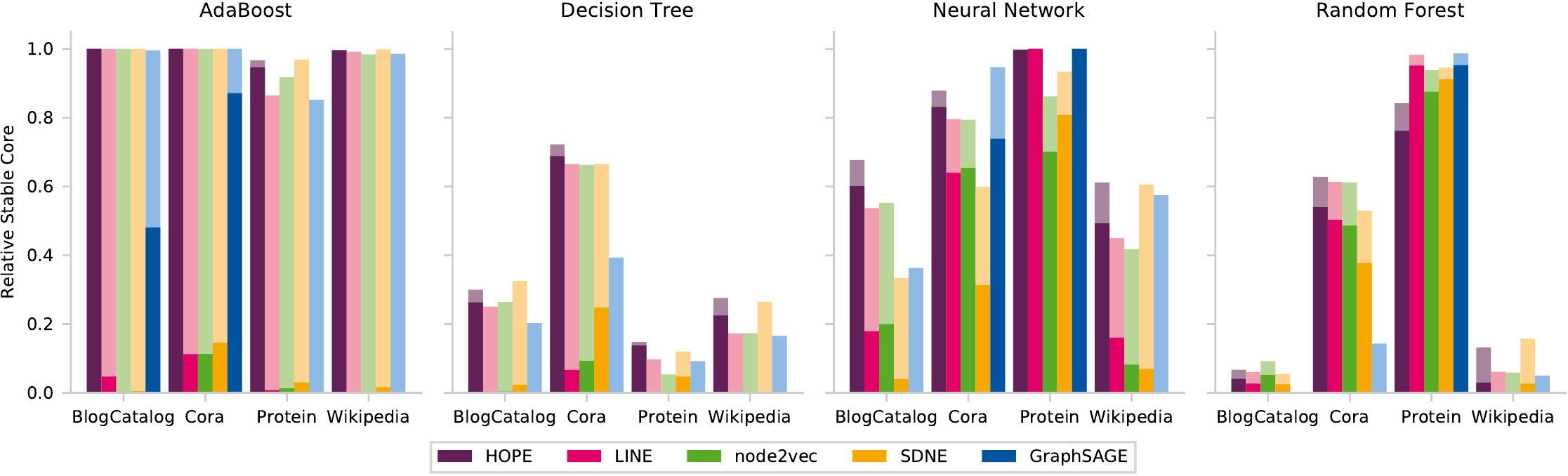}
	\caption{\emph{Stability in node-wise predictions.} Stability of the classifiers as ratios of nodes which are always predicted to be in the same class. Saturated colors represent the mean stable core of all 30 embeddings and lighter colors the mean stable core of five randomly sampled embeddings with 10 repetitions each. The stability of node classification heavily depends on the dataset and embedding algorithm. AdaBoost is good in reproducing the classification over repetitions.
	Random forests are mainly unstable due to inherent instability, independent of the underlying embedding.}
	\label{fig:overlap}
\end{figure*}
We now investigate the stability of the \emph{node-wise predictions} themselves.
Thus, on each dataset, we compute the stable core on the test set over different instances of the same classifier.
The instances of the classifier were trained 
\begin{enumerate}[(i)]
    \item ten times on five randomly sampled embeddings,
    \item once on each of the $30$ embeddings per algorithm and dataset.
\end{enumerate}
In Figure \ref{fig:overlap}, we visualize both parts of the experiment, the stable core for classifiers trained on the same embedding (averaged over five embeddings) in light colors (i) and the stable core for the classifiers trained on different embeddings in saturated colors~(ii).
In case that both bars are of the same height, the largest part of the instability is due to the inherent instability of the classifier and there is no significant influence of the embedding.
Otherwise, if the the stable core from (i) is significantly lower than the stable core from (ii), the random variations of the embeddings have a significant influence on the stability of the classification.

Our first observation is that since the embeddings generated by HOPE are almost identical, also the ratio in (i) and (ii) is about the same over all datasets and classifiers. Next, we compare the stability of the different classifiers on the other embedding algorithms.
For AdaBoost, we observe that the stable core (ii) is always lower than \(0.05\), except for the Cora graph and GraphSAGE on BlogCatalog.
Thus, almost all nodes are assigned different labels when identical classifiers are trained on different embeddings created by the same algorithm.
In contrast, AdaBoost is nearly deterministic on fixed embeddings (i).
Random forests are the exact opposite to AdaBoost regarding this measure. 
Here, for most embedding algorithms and datasets the stable cores from (i) and (ii) match, indicating that only the inherent instability of the classification algorithm influences the stability of the predictions.
For decision trees, we see rather small stable core in (ii) which drops even further in (i), indicating that due to the randomized implementation, decision trees output highly unstable classifications. 
This instability is further amplified when training on different embeddings, especially GraphSAGE embeddings.
Only on Cora the stable core in~(ii) reaches values of around \(0.7\). 
For neural networks, we observe stable predictions on Protein over all embedding methods, and almost as stable predictions for Cora, while on Wikipedia the stable core (ii) is minimal.

Finally, we list a few more notable observations, which are not covered by the general descriptions.
On the Protein dataset, random forests and neural networks produce extremely stable predictions while achieving less than 5\% accuracy, \ie{} the classifier consistently predicts a wrong answer.
In neural network predictions, LINE and node2vec embeddings allow for more stable predictions than SDNE.

\section{Discussion}\label{sec:discussion}

This section provides a short summary of results, links the observed instability of algorithms to sources of randomness they contain, and discusses potential implications and limitations of our work.

\subheader{Summary of results. }
Our results indicate clear differences in the geometric stability between the embedding algorithms.
HOPE consistently yields near-constant embeddings.
In contrast, GraphSAGE can be considered as very volatile. 
The other algorithms (LINE, node2vec and SDNE) also exhibit a substantial degree of instability---for example, when running the same embedding algorithm twice over the same data, 20\% to 80\% of the nearest neighbors of each node match strongly influenced by the dataset.
The influence of graph properties on node embedding stability is highly dependent on the embedding algorithm. 
While GraphSAGE shows decreasing stability for larger graphs, node2vec and SDNE tend to increase stability on more dense and larger graphs.

Regarding downstream node classification, our results show that the overall classification accuracy is mostly unaffected by random variations in the embeddings.
However, the actual predicted classes for single nodes vary depending on the embedding the classifier was trained on, \ie{} due to the randomness in the embeddings, different classification errors are made.
This effect strongly depends on the employed classification algorithm: decision trees and AdaBoost seem to be most influenced by this effect, while neural networks and random forests are much more robust against instabilities of the embeddings.

\subheader{Potential explanations for results. }
The different degrees of embedding stability correspond in general to the inherent sources of randomness in the embedding algorithms.
First of all, HOPE achieves near-constant embeddings since the only source of randomness is the approximation of the SVD, which converges very well most of the time.
In contrast, GraphSAGE computes its embeddings by aggregating features of randomly sampled neighborhoods.
Although the aggregation function is optimized through learning, this process is inherently unstable.

Regarding the other three algorithms, the combination of low $k$-NN Jaccard similarity and high second-order cosine similarity could imply a clustering of the embedding space.
Definitely we see that similar nodes are mapped close together in the embedding space.
SDNE's local loss function actively encourages similar nodes to be mapped close together,
while the global loss function is not explicitly designed to map distant nodes far away in the embedding space.
Thus, especially for this embedding algorithm, clusters in the embedding space are likely.
In node2vec, nodes which appear in a similar neighborhood are embedded to aligned vectors since their random walks have large common subsets.
This again supports groups of nodes to appear close together.
A similar argument can be made for LINE, although this algorithm's loss function rather focuses on keeping one-hop or two-hop neighborhoods together.
This focus also explains why for LINE the angles between the embeddings of neighboring nodes vary less than for more distant nodes, although this effect is rather small.

Regarding the stability of classification performance, a potential explanation is that classifiers are able to extract and utilize local structural information from embeddings even if the embedding's global structure changes.
This results in almost constant classification accuracies even with unstable embeddings.

\subheader{Implications.}
In the authors' opinion, the outcomes of this paper have significant impact on the research of node embeddings.
Since node embeddings can vary just based on their internal random processes, great care must be taken in their evaluation and, if possible, experiments should be repeated several times in order to estimate and limit the influence of randomness and enable reproducibility of results. 
This is specifically important if geometric properties such as the local neighborhood are exploited. 

Node embedding algorithms are often applied to social network data where nodes represent real people. 
Especially when high-stakes real-world decisions take into account techniques involving node embeddings, practitioners should be aware that the use of node embeddings adds another level of uncertainty into the individual (e.g., classification) decision.
An interesting little investigated question in node embedding research is also the role of privacy, cf.~\cite{ellers2019privacy}.
When it comes to typically large social networks, retraining of node embeddings for every change (e.g., deletion of a user node) is expensive.
At the same time the privacy of the users has to be guaranteed upon deletion, but might be in danger due to information contained in the embeddings of other nodes.
From that point of view, instability of embeddings might not only be a challenge, but a useful feature that complicates recovery of deleted nodes and thus supports user privacy. 
In that regard, our work constitutes a first step towards understanding the role of node embedding stability.

\subheader{Limitations. }
The stability of the investigated algorithms might be strongly influenced by their concrete implementations. In that regard, we picked reference implementations from the respective research papers or---if that was not possible---established code bases for the different algorithms. 
However, we cannot rule out that some (in-)stabilities we observed are a consequence of implementation details. 
Since the chosen implementations are widely used, our results are still highly relevant for researchers and practitioners.

As an empirical study, our work does not contribute directly to identifying sources of algorithmic instability in a principled manner. 
In that regard, a theoretical analysis of the individual algorithms could be an interesting direction of future work.

Additionally, our study is currently limited to small and medium sized graphs. 
This is due to the computational requirements of the repeated  calculation of node embeddings. 
However, our study of the influence of graph properties, see Section~\ref{subsec:geometricStabilityResults}, already provides a clear indication of trends that might manifest in larger graphs.

\section{Related Work}\label{sec:relatedWork}
Research on word embedding algorithms has hugely influenced the research on node embeddings, including our work. Thus, we give an overview on related work on the stability of both word and node embeddings.

\subheader{Stability of Word Embeddings.}
	There are a number of publications that investigate the stability of word embeddings, a topic that gained a lot of attention when word embeddings were applied to investigate semantic differences in language in historical or cultural context.
When studying such semantic aspects, \mbox{\citet{hellrich2016bad}} discovered that neighborhoods of words in the embedding space change significantly even under fixed corpora.
	These instabilities have been confirmed and further investigated by \citet{antoniak2018stability}. 
	Both studies \cite{hellrich2016bad, antoniak2018stability} report significant instabilities of skip-gram-based word embedding methods with respect to local neighborhood similarities. Their results coincide with ours on skip-gram based methods such as LINE and node2vec.
	To investigate which word properties influence this instability, \citet{wendlandt2018instability} and \citet{pierrejean2018predicting} conducted regression analyses to determine factors that contribute to the instability of word embeddings.

Regarding mathematical properties of word embeddings, \citet{levy2014neural} and \citet{levy2015improving} point out that skip-gram based word embeddings implicitly factor a shifted word-context PMI matrix, whose performance is highly dependent on the choice of hyperparameters.
    \citet{mimno2017geometry} investigated geometric properties of the embedding spaces of such skip-gram based word embedding algorithms.
    
\subheader{Stability of Node Embeddings.} 
In contrast to the situation for word embeddings, there are only few publications on the stability and geometry of node embeddings.
In two corresponding survey and benchmark studies, \citet{goyal2018survey} and \mbox{\citet{goyal2019benchmark}} experimentally tested the downstream performance of numerous embedding algorithms under varying parameters and graph properties. 
  Their results indicate that the performance in node classification and link prediction was often strongly affected by parameter choice, particularly for HOPE and SDNE.
  For varying graph density, there appeared to be no clear effect on downstream performance, while for larger graph sizes accuracies tended to deteriorate.

  In a publication that can be seen as an adaptation of the work by \citet{levy2014neural}, \citet{qiu2018network} show that skip-gram-inspired embedding algorithms like DeepWalk, LINE and node2vec all perform implicit matrix factorizations.
  Based on that, they present an algorithmic framework to factorize such matrices.
More recently, \citet{liu2019general} have shown that matrix factorization approaches are equivalent to approaches simultaneously mapping similar nodes closely together and separating distant nodes.

  Finally, the issue of robustness of graph convolutional network embeddings against adversarial attacks is tackled in a recent series of publications by \citet{zugner2018adversarial} and \citet{zugner2019adversarial, zugner2019certifiable}. 
  In particular, the authors show that the performance of node classification can significantly deteriorate under small perturbations of the input graph \cite{zugner2018adversarial, zugner2019adversarial}, and have developed a method to certify (non-)robustness of specific nodes to such attacks \cite{zugner2019certifiable}.

\section{Conclusion}
    In this work, we found substantial variability between node embeddings from different runs of the same algorithm with the same parameters and input data with respect to the geometry in the embedding space.
    Considering node classification as an example for downstream tasks, we found only small variations with respect to the overall classification accuracy, but considerable variations between the classifications of individual nodes.

    In the future, we anticipate investigations of stability and robustness of node embedding algorithms towards an in-depth study of the effects of different embedding sizes and graph modifications such as deletions or additions of nodes, subpopulations of nodes, or edges.
    A better understanding of the impact of graph modifications on embeddings is gaining importance given new regulations such as the European GDPR that require companies to respond to requests for deletion by individuals. 
    As a consequence, it may be desirable to design embedding algorithms that control the variability of an embedding.
    Such algorithms might allow to reliably delete information about nodes from an embedding without recomputing the embedding and re-running subsequent machine learning algorithms.

    Furthermore, we see an opportunity for developing measures
that will allow to estimate the potential instability of an embedding without computing it multiple times.
    Another interesting angle could be to examine stability of node embedding from the perspective of learning theory, which asks how much a small perturbation in the input can modify a learned classifier in theory.

\section*{Acknowledgements}
This work is supported by the German research council (DFG) Research Training Group 2236 UnRAVeL

\bibliographystyle{plainnat}

\end{document}